\documentclass{article}
\usepackage{spconf,amsmath,graphicx,times,epstopdf,subfig,float,siunitx}

\usepackage{hyperref,xcolor}
\hypersetup{
    colorlinks,
    linkcolor={red!50!black},
    citecolor={blue!50!black},
    urlcolor={blue!80!black}
}

\DeclareSIUnit\px{px}

\title{REFLECTION SEPARATION USING GUIDED ANNOTATION}
\name{Ofer Springer\,\textsuperscript{*}\thanks{\textsuperscript{*}springer@cs.huji.ac.il}\textsuperscript{\textdagger}, Yair Weiss\,\textsuperscript{\textdagger}\thanks{\textsuperscript{\textdagger}Supported by Intel ICRI-CI and the ISF.}}
\address{School of Computer Science and Engineering\\
  The Hebrew University of Jerusalem\\
  91904 Jerusalem, Israel}

\begin{document}

\maketitle

\begin{abstract}
Photographs taken through a glass surface often contain an approximately linear superposition of reflected and transmitted
layers. Decomposing an image into these layers is generally an ill-posed task and the use of an additional image prior and user provided cues is presently necessary in order to obtain good results. Current annotation approaches rely on a strong sparsity assumption. For images with significant texture this assumption does not typically hold, thus rendering the annotation process unviable.

In this paper we show that using a Gaussian Mixture Model patch prior, the correct local decomposition can almost always be found as one of 100 likely modes of the posterior. Thus, the user need only choose one of these modes in a sparse set of patches and the decomposition may then be completed automatically. We demonstrate the performance of our method using synthesized and real reflection images.
\end{abstract}


\section{Introduction}
\label{sec:intro}

Many real-world photographs contain reflections and photographers often expend significant effort to avoid their effect. Ideally, we would like a fully automatic method that can  remove reflections in post-processing of a single image. In the general single image reflection separation problem, assuming a linear response of the camera sensor then the input image $y=x_1+x_2$ is a sum of two unknown reflection and transmission images. Automatically recovering $x_1$ and $x_2$ given $y$ is a highly ill-posed task. This is due to the fact that the number of equations is generally half the number of unknowns. 

The problem becomes less difficult when additional cues exist. Some existing approaches use the presence of a double image \cite{ShihKrishnan} or polarization \cite{FaridAdelson,SchechnerShamir} in the reflection layer. Others use the availability of more than a single composite image due to camera motion \cite{szeliski2000layer,gai2012blind,XueRubinstein} or additional a priori known differences between the reflected and transmitted layers, e.g. a smoothness disparity \cite{li2014single} or a disparity in the relative layer attenuation factor \cite{yan2013separation}. However, in many situations, these extra cues are not available.

While fully automatic separation of reflection images from a single image is extremely difficult, photographers would also be happy with a {\em semi-automatic} method requiring user annotation. If the amount of annotation required is manageable then a method that completes the separation given this annotation may be of great utility. This was the motivation behind the work of Levin and Weiss \cite{LevinWeiss} (LW) who presented a user-assisted method for separating reflections. Their method was based on the well-known property that derivative filters of natural images tend to have a sparse distribution. This property both served as an image prior and was the basis of their proposed annotation mechanism: the user was tasked with selecting pixels in which the (first and second order image derivative) filter responses fully originated from only one of the layers. They then optimized for a likely reflection separation that is also consistent with these annotations.

The LW method is the only currently available approach to solving the single image user-assisted reflection separation task that we know of. As shown in \cite{LevinWeiss} it is often possible to get very good separations on real images with a modest amount of user input. However, for many images, we find that the annotation mechanism of the LW method is not applicable. Often a large fraction of the pixels in one layer contain non-sparse texture overlapping edges in the other layer. In such cases it is often impossible to locate a sufficient number of pixels having filter responses originating from only a single layer. 
\begin{figure}[h]

\centerline{\includegraphics[width=0.35\paperwidth]{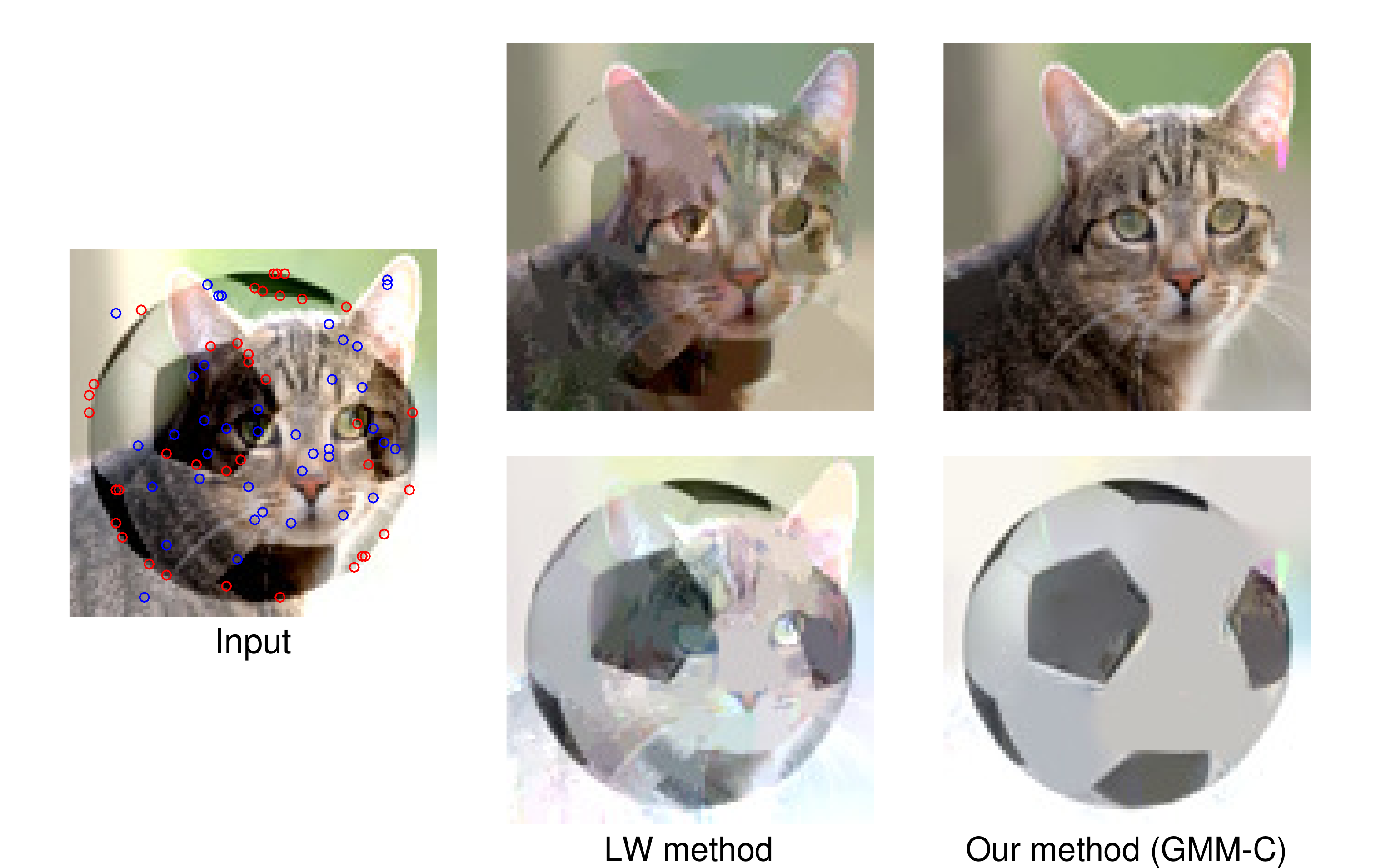}}
\caption[]{Visual comparison of the results of running the sparsity based LW method and our method on a synthetic reflection image.}
\label{fig:teaser}
\end{figure}
Figure \ref{fig:teaser} shows one such example. The input image is a sum of two images, the first of which contains significant texture. Due to this texture, filter responses in both images are nonzero at edges of the second image, and thus the LW labeling mechanism fails. We show in figure \ref{fig:teaser} the results of running the LW method and our method on this input.

In section \ref{sec:sparsity} we study the limitations of the sparsity assumption as an annotation mechanism. In section \ref{sec:gmm} we introduce our use of a GMM patch prior and derive the patch posterior for the reflection separation task in section \ref{sec:posterior}. An alternative annotation mechanism that utilizes this posterior and overcomes the limitations of the LW annotation mechanism is presented in section \ref{sec:annot}. The full algorithm utilizing both the revised prior and revised annotation mechanism (GMM-C) is described in section \ref{sec:alg}. In section \ref{sec:results} we asses the accuracy of our proposed method on reflection images synthesized from the BSDS300 dataset \cite{MartinFTM01} and show separation results on real reflection images.

\section{Limitations of the sparsity assumption}
\label{sec:sparsity}
\begin{figure}[h]
\centerline{
\begin{tabular}{ccc}
\subfloat{\includegraphics[width=0.25\columnwidth]{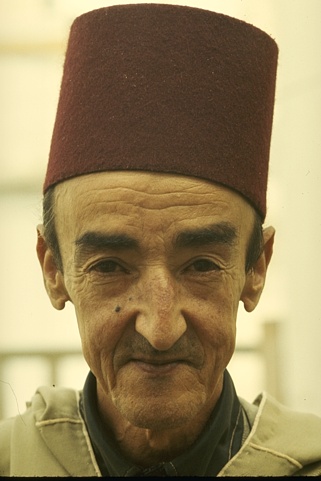}}\hspace{0.05em}
\subfloat{\includegraphics[width=0.25\columnwidth]{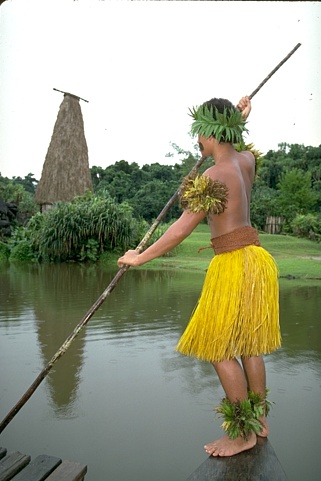}}\hspace{0.05em}
\subfloat{\includegraphics[width=0.25\columnwidth]{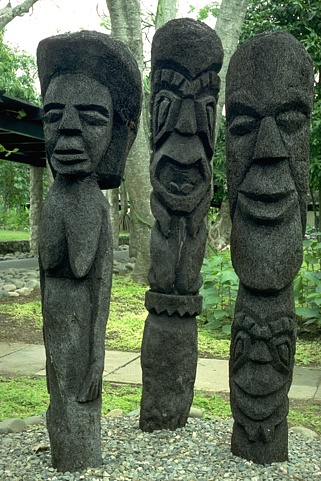}}\\[-2.3ex]
\subfloat{\includegraphics[width=0.25\columnwidth]{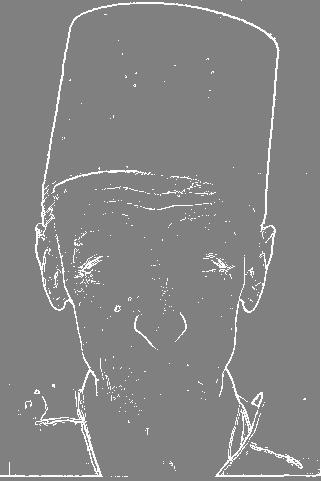}}\hspace{0.05em}
\subfloat{\includegraphics[width=0.25\columnwidth]{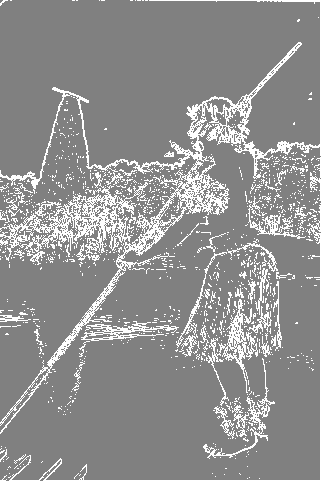}}\hspace{0.05em}
\subfloat{\includegraphics[width=0.25\columnwidth]{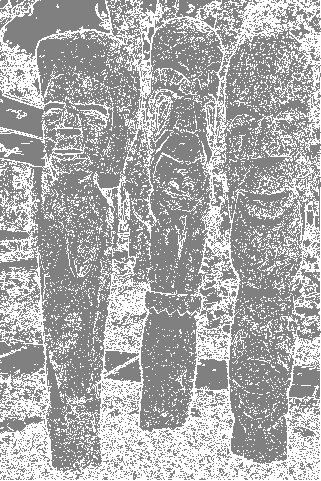}}\\[-2.8ex]\hspace{-2em}
\subfloat{\includegraphics[width=0.80\columnwidth]{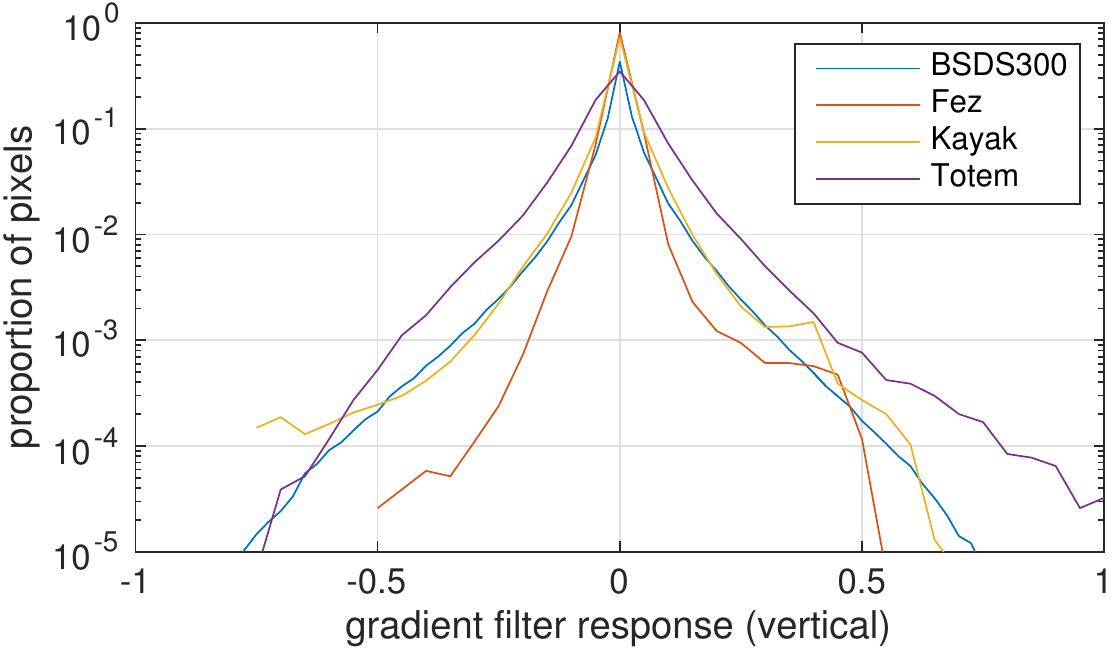}}
\end{tabular}}
\caption{Example images having a large variation in filter response statistics shown in the top row with corresponding pixels having gradient magnitude $>0.1$ marked white in the middle row.  We see that the fraction of high gradient magnitude pixels varies greatly and that vertical gradient histograms also display this over and under sparseness of the ``Fez'' and ``Totem'' images compared to the ``Kayak'' image and the full BSDS300 dataset.}
\label{fig:stats_filts_examples}
\end{figure}

Although the outputs of derivative filters applied to natural images tend to have a sparse distribution overall (c.f. \cite{Hyvarinen}), this property varies greatly between images and within images. An example of this is shown in figure \ref{fig:stats_filts_examples}, where three natural images taken from the BSDS300 dataset show very different filter response statistics.
We can also quantify this disparity between natural images by noting that although overall in BSDS300 ${\sim}10\%$ of pixels have gradient magnitude $> 0.1$, it is also the case that ${\sim}30\%$ of the images in BSDS300 have ${\sim}30\%$ of pixel gradient magnitudes $> 0.1$. This indicates that this sparsity property is not spread out evenly. Some images tend to contain much more textured regions than others and filter responses in these textured regions tend to have a non-sparse gaussian distribution. 

\begin{figure}[h]
\centerline{
\begin{tabular}{cccc}
\subfloat{\rotatebox{90}{\footnotesize{\hspace{1.60em}GT\hspace{1.65em}GMM-C\hspace{1.60em}LW\hspace{2.30em}Input}}}
\subfloat{\includegraphics[width=0.28\columnwidth]{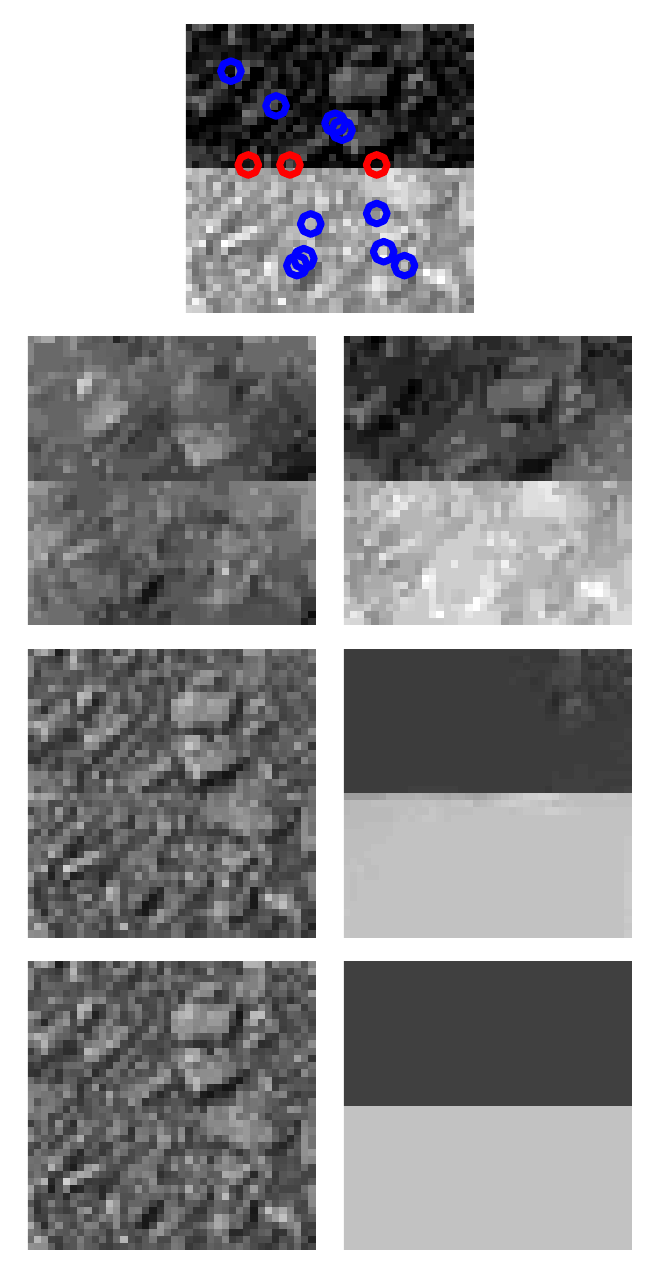}}\hspace{0.2em}\rule{0.5px}{133px}
\subfloat{\includegraphics[width=0.28\columnwidth]{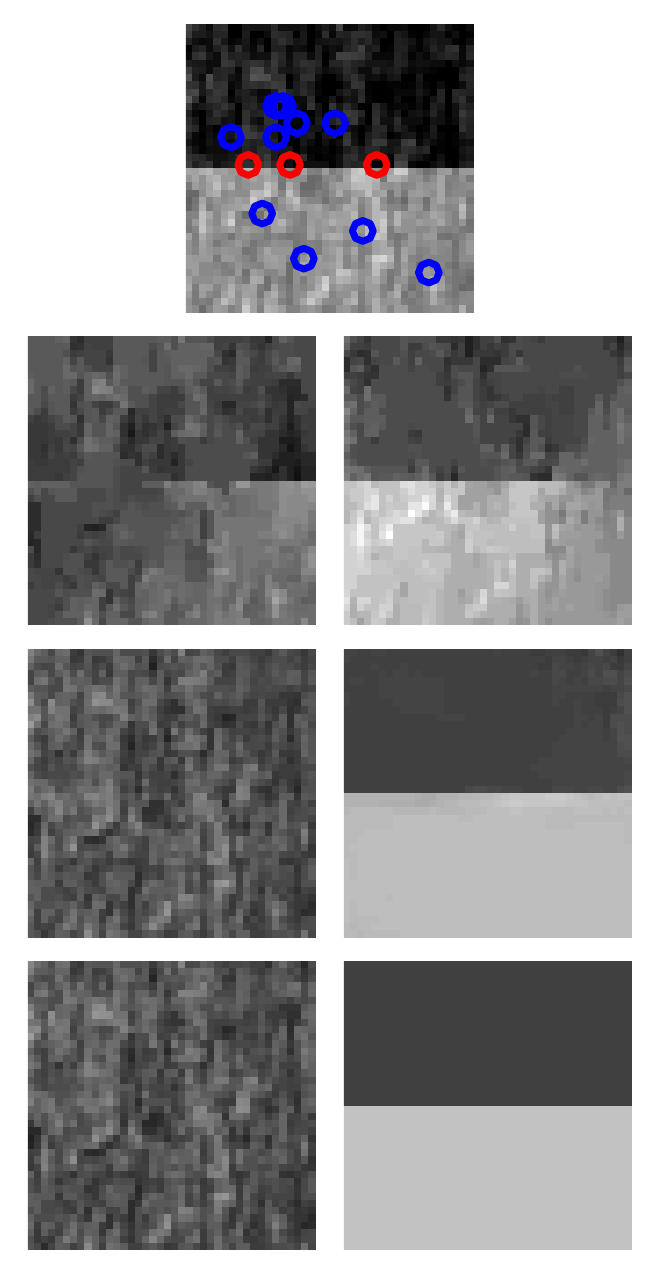}}\hspace{0.2em}\rule{0.5px}{133px}
\subfloat{\includegraphics[width=0.28\columnwidth]{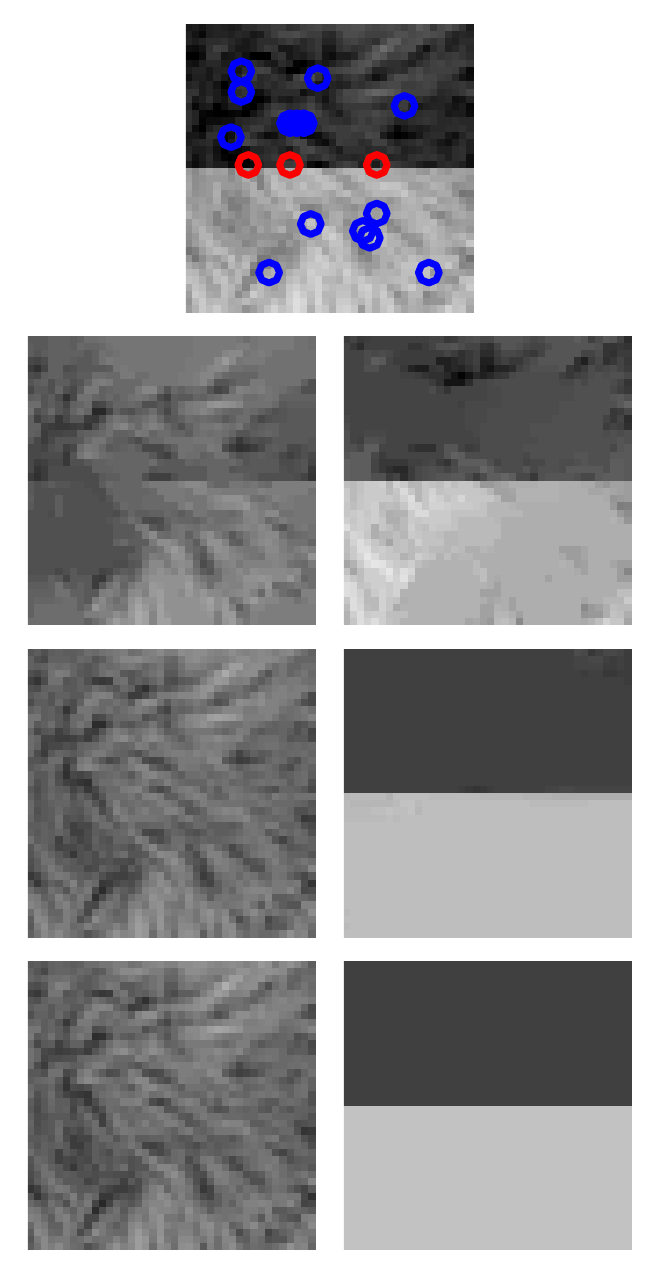}}
\end{tabular}}
\caption{Visual comparison of decomposing non sparse textures (asphalt, tree bark, fur) from a horizontal edge using the LW and GMM-C methods. Bottom row shows ground truth.}
\label{fig:edge_texture}
\end{figure}

When one layer contains texture, it becomes difficult or oftentimes impossible to correctly annotate edges in the overlapping regions of the second layer using the LW annotation mechanism. In figure \ref{fig:edge_texture}, we show how these unavoidable inaccuracies in annotation affect the quality of the resulting LW separation.
In all cases, annotation is performed automatically using the ground truth layers and the automatic annotation protocol of \cite{LevinWeiss}. In general we see in the LW results that the inaccurate annotation leads to texture transferring into the edge layer and vice versa. In section \ref{sec:results} we quantify this visual observation.

\section{The GMM Patch Prior}
\label{sec:gmm}
While the sparsity of filter responses may often serve as a reasonable prior, recent work has shown that more powerful models of natural images can be learnt. One very successful prior is the Gaussian Mixture Model (GMM) that models the statistics of $8 \times 8$ image patches (see \cite{ZoranWeiss}). Denoting the 64 dimensional vector representation of the patch by $x$, then under the GMM natural image patch prior we have
\[
P_{nat}(x) = \sum_{k=1}^K \pi_k \mathcal{N}(x;\mu_k,\Sigma_k),
\]
where $\pi_k$ is the mixture weight of component $k$ and $\mathcal{N}(x;\mu_k,\Sigma_k)$ denotes a Gaussian density with mean $\mu_k$ and full covariance matrix $\Sigma_k$. In a similar fashion to \cite{ZoranWeiss} we learn two models having $K=50$ and $K=200$ components using the BSDS300 train dataset. 

\section{The Patch Posterior}
\label{sec:posterior}
Suppose we observe an image patch $y=x_1+x_2$ that is the sum of two i.i.d. natural image patches $x_1, x_2 \sim P_{nat}(x)$, then the posterior probability of observing $x_1$ given $y$ is
\begin{equation*}
\label{eq:posterior}
\begin{split}
\Pr(x_1|y) &= \int dx_2\,\Pr(x_1,x_2|y)\\
&= \int dx_2\,\Pr(y|x_1,x_2)\Pr(x_1,x_2)/\Pr(y) \\
&= \int dx_2\,\delta(x_1+x_2-y)\Pr(x_1)\Pr(x_2)/\Pr(y) \\
&= \frac{1}{Z}P_{nat}(x_1)P_{nat}(y-x_1).
\end{split}
\end{equation*}
Expanding each $P_{nat}(\makebox[1ex]{\textbf{$\cdot$}})$ factor we obtain the following expression for the posterior
\begin{equation*}
\label{eq:posterior_expanded}
\begin{split}
\Pr(x_1|y) &= \frac{1}{Z}
\sum_{i=1}^K\sum_{j=1}^K\pi_i\pi_j \mathcal{N}(x_1;\mu_i,\Sigma_i)\mathcal{N}(y-x_1;\mu_j,\Sigma_j) \\
&= \frac{1}{Z}
\sum_{i=1}^K\sum_{j=1}^K\pi_i\pi_j \mathcal{N}(x_1;\mu_i,\Sigma_i)\mathcal{N}(x_1;y-\mu_j,\Sigma_j), \\
\end{split}
\end{equation*}
and using a property of products of Gaussian densities \cite{Petersen}, this can be written as a GMM having $K^2$ components, one for each Gaussian cross term
\begin{equation*}
\label{eq:square_gmm}
\begin{split}
\Pr(x_1|y) = \sum_{i=1}^K\sum_{j=1}^K\pi_{ij}\mathcal{N}(x_1;\mu_{ij},\Sigma_{ij}),
\end{split}
\end{equation*}
where
\begin{equation*}
\label{eq:square_gmm}
\begin{split}
\Sigma_{ij} &= \left(\Sigma_i^{-1}+\Sigma_j^{-1}\right)^{-1} \\
\mu_{ij}(y) &= \Sigma_{ij}\left[\Sigma_i^{-1}\mu_i+\Sigma_j^{-1}(y-\mu_j)\right] \\
\pi_{ij}(y) &= \frac{1}{Z} \pi_i\pi_j \mathcal{N}(\mu_i+\mu_j;y,\Sigma_i+\Sigma_j).
\end{split}
\end{equation*}
Note that the mixture weights and means of the posterior GMM depend on the prior GMM as well as on the input patch $y$, while the posterior covariances depend only on the prior covariances and may be precomputed. We also note that the resulting posterior means will generally be nonzero. Given an input image patch $y$ a naive approach to finding the original patches is to find $x_1$ that maximizes $\Pr(x_1|y)$. Unfortunately due to the posterior probability being highly multimodal such an approach often fails. A more potent approach is to rely on the user at this point and ask that he picks $x_1$ from a set of candidate decompositions that would likely contain a close match to the true decomposition. We therefore seek for a set of preferably diverse decompositions that maximize $\Pr(x_1|y)$.

\section{A new annotation method}
\label{sec:annot}
We choose to approximate such a set of candidate decompositions by taking the means of the posterior GMM that have highest mixture weights. 
\begin{figure}[htb]
\centerline{
\begin{tabular}{c}
\subfloat{\includegraphics[width=0.99\columnwidth]{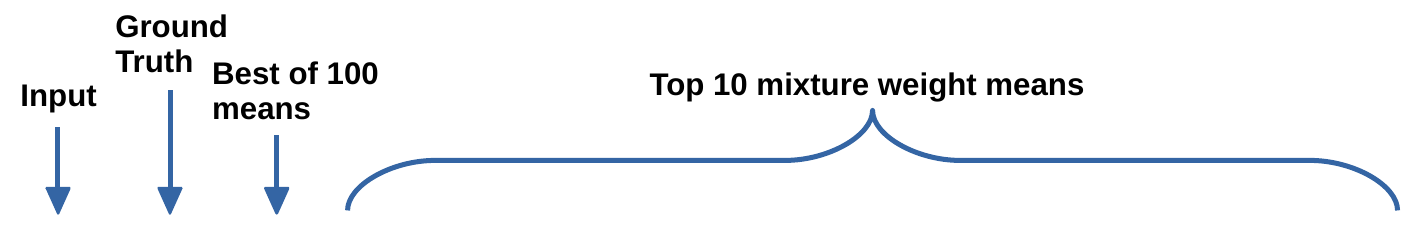}}\\[-3.5ex]
\subfloat{\includegraphics[width=0.99\columnwidth]{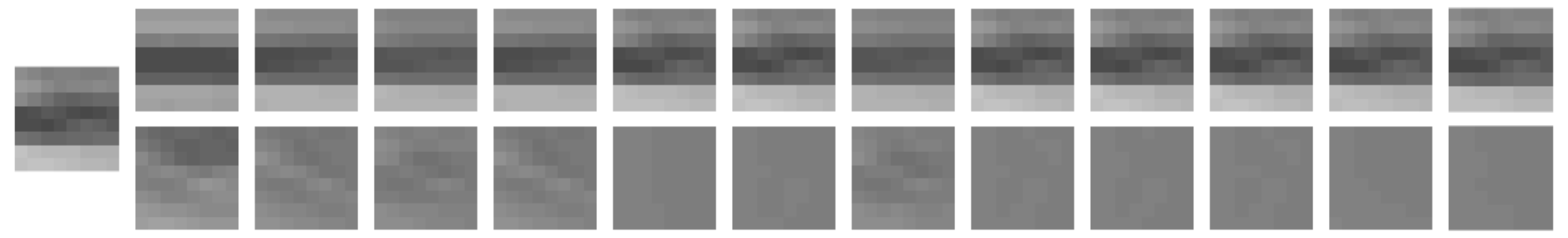}}\\[-2.1ex]\rotatebox{90}{\rule{0.5px}{240px}}\\[-2.6ex]
\subfloat{\includegraphics[width=0.99\columnwidth]{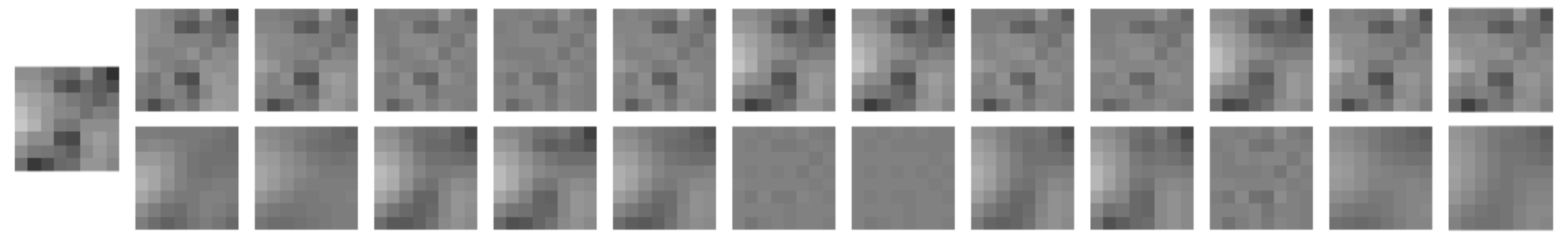}}\\[-2.1ex]\rotatebox{90}{\rule{0.5px}{240px}}\\[-2.6ex]
\subfloat{\includegraphics[width=0.99\columnwidth]{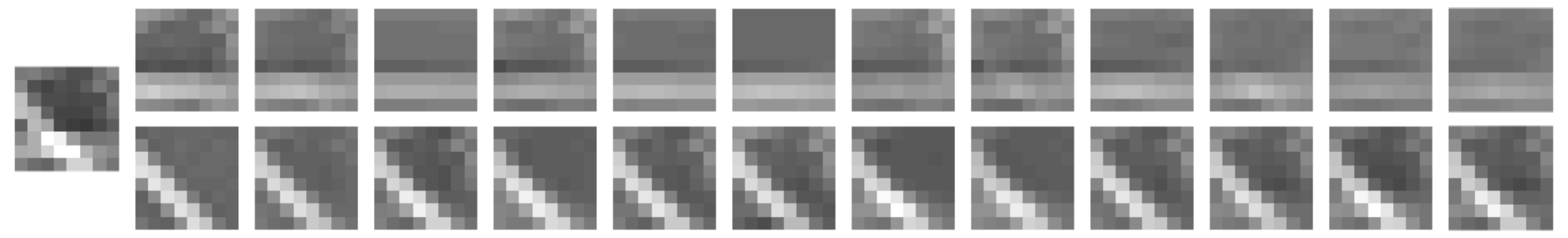}}
\end{tabular}}
\caption{Some examples of $8\times 8$ patch decomposition candidates provided by the posterior means having top mixture weights. Here the $K=50$ prior was used.}
\label{fig:patch_decomps}
\end{figure}
Figure \ref{fig:patch_decomps} shows some examples of these local decompositions. Even when it is hard for an unguided human to decompose $y$ into the original patch layers, the posterior GMM does a remarkable job of suggesting good candidates among the top $100$ posterior means and often even among the top $10$. 

\begin{figure}
\centerline{
  \includegraphics[width=0.85\columnwidth]{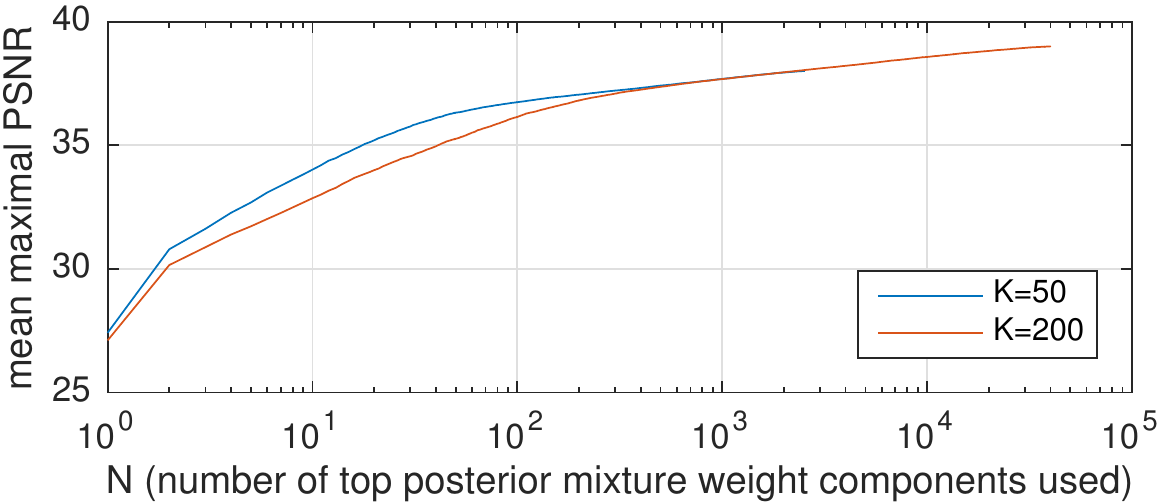}}
\caption{Accuracy of best patch decomposition among $N$ candidates: the $N$ posterior means with highest mixture weights.}
\label{fig:stats_annot_comps}
\end{figure}

To quantify the performance of this candidate proposal mechanism we repeatedly sampled a random pair of image patches $x_1, x_2$ from the BSDS300 test set, and summed them to create $y=x_1+x_2$. Given $y$ we created a set of $N$ possible decompositions by taking the $N$ posterior means with highest mixture weights, and measured the distance between the true $x_1$ and the closest posterior mean among these $N$ candidates. Figure \ref{fig:stats_annot_comps} shows that  with as few as $N=100$ candidates, it is possible to find a decomposition that is a very close match to the true $x_1$ (with accuracies of $\SI{36.7}{\decibel}$ for $K=50$ and $\SI{36.2}{\decibel}$ for $K=200$).

Our proposed annotation method is then to require the user to both pick an annotation point and to pick the most appropriate patch decomposition among the $N=100$ posterior means with highest mixture weights at this point.

\section{Full algorithm}
\label{sec:alg}
Given input image $y$ which is the sum of two unknown layer images $x_1,x_2$ the user is first prompted to provide annotation at a set of image points as follows: 
\begin{itemize}
\item The user picks an annotation point $i^*$ and the surrounding input patch $P_{i^*}y$ is extracted. Here $P_{i^*}$ is the linear operator extracting an $8\times 8$ pixel patch from an image encoded as a column vector.
\item Given $P_{i^*}y$, the corresponding set of $N = 100$ posterior means with highest mixture weights are computed and presented to the user as candidate decompositions.
\item The user then picks the decomposition most appropriate at point $i^*$. We denote by $\mu_{i^*}$ the posterior mean selected by the user and the corresponding posterior covariance by $\Sigma_{i^*}$.
\end{itemize}
The annotation is then automatically propagated by finding image $x_1$ maximizing the expected patch log likelihood (EPLL) function
\[
EPLL(x_1|y) = \sum_i \log\Pr( P_i x_1| P_i y),
\]
under the set of annotation constraints $P_{i^*}x_1 = \mu_{i^*}$. Here $i$ indexes all overlapping image patches and $i^*$ indexes patches surrounding annotation points. In practice we replace these hard constraints by an additional cost and minimize
\begin{multline*}
J_C(x_1)= - EPLL(x_1|y) \\
+ \frac{\lambda_C}{2} \sum_{i^*} (P_{i^*} x_1- \mu_{i^*})^T \Sigma_{i^*}^{-1} (P_{i^*} x_1 - \mu_{i^*}).
\end{multline*}
This is the cost used in our proposed GMM-C method (GMM with component annotations). To optimize this cost we use the iterated least squares approach described in \cite{LeviWeiss}. Once $x_1$ is estimated, the second image is then set to $x_2 = y-x_1$ per the problem definition.

\section{Results}
\label{sec:results}

\begin{figure}
\centering
\centerline{\includegraphics[width=0.85\columnwidth]{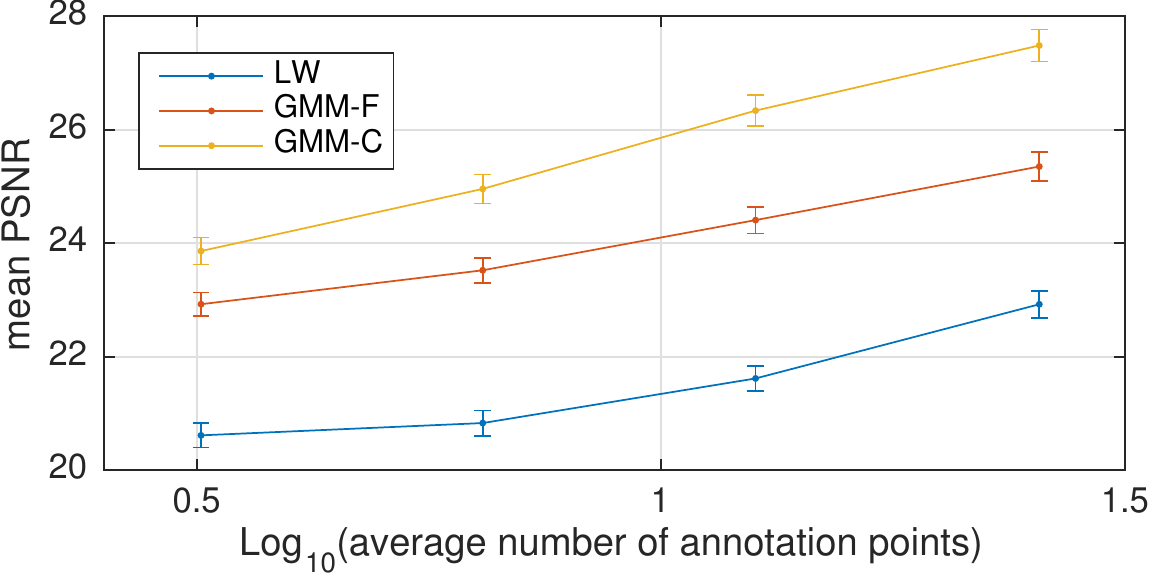}}
\caption{Decomposition accuracy of LW, GMM-F and GMM-C measured on $8,\hspace{-0.13em}000$ $40\times40$ pixel patches generated from random BSDS300 test patch pairs.}
\label{fig:stats_accuracy}
\end{figure}

Accuracy was assessed by decomposing $8,\hspace{-0.13em}000$ $40\times 40$ pixel patches generated from the addition of random BSDS300 test patches. To separately measure the effects the revised prior and revised annotation each have on the relative accuracies of the GMM-C and LW methods, we also included a third control method denoted GMM-F (GMM with filter annotations). GMM-F uses the annotation cost term of the LW method (see \cite{LevinWeiss}) but replaces the prior cost term with the negative EPLL cost defined in the previous section. 

Filter annotations for the LW and GMM-F methods were generated automatically using the protocol used by \cite{LevinWeiss}. Essentially a small random subset of the canny edges in the input image were classified as originating from either layer, depending on which ground truth layer had greater gradient magnitude there. Automatic annotation for GMM-C was computed at this same set of locations. At these locations one of the top two posterior means closest to the ground truth among the set of $N=100$ posterior means with highest mixture weights was randomly picked. All results presented in this section were generated using the $K=200$ GMM prior (both in the auto-annotation protocol and in the EPLL cost). 

Accuracy statistics are presented in figure \ref{fig:stats_accuracy}. The overall PSNR gain between the LW and GMM-C methods ranges between $\SI{3.3}{\decibel}$ at an annotation density of $1/{\SI{22}{\px}^{-2}}$ and $\SI{4.6}{\decibel}$ at $1/{\SI{8}{\px}^{-2}}$. Of this gain, approximately $\SI{2.5}{\decibel}$ is due to the modified prior (GMM-F) and approximately $\SI{1.6}{\decibel}$ is due to the modified annotation. 

\begin{figure}[htb]
\centerline{
\begin{tabular}{ccccc}
\subfloat{\rotatebox{90}{\footnotesize{\hspace{3.5em}Input}}}\hspace{0.0002em}
\subfloat{\includegraphics[width=0.265\columnwidth]{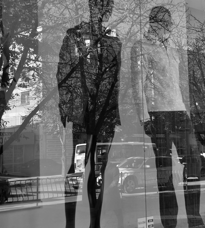}}\hspace{-0.06em}
\subfloat{\includegraphics[width=0.265\columnwidth]{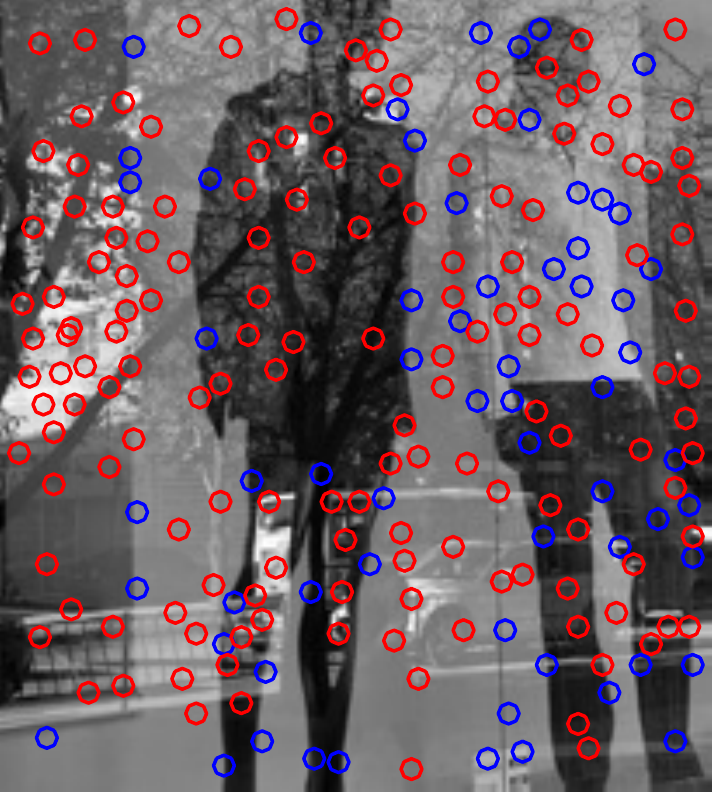}}\hspace{-0.06em}
\subfloat{\includegraphics[width=0.2135\columnwidth]{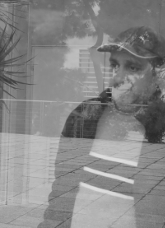}}\hspace{-0.06em}
\subfloat{\includegraphics[width=0.2135\columnwidth]{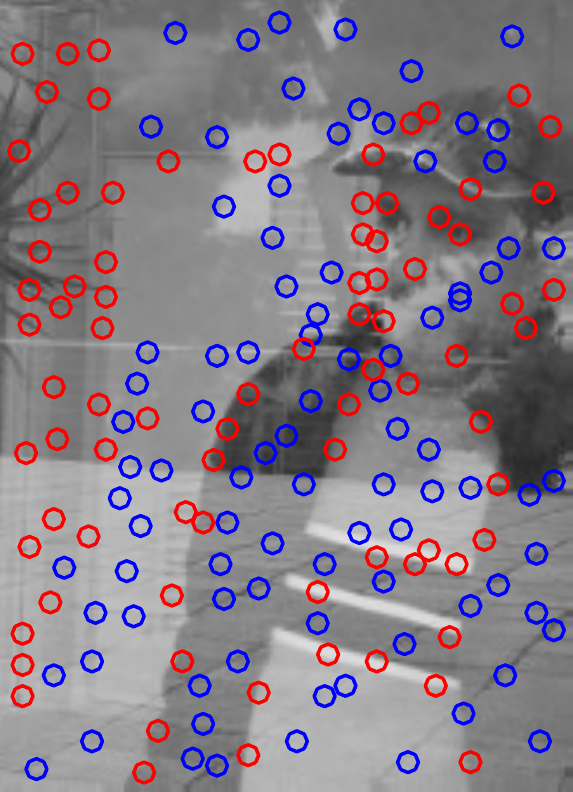}}\\[-2.5ex]
\subfloat{\rotatebox{90}{\footnotesize{\hspace{3.5em}Layer 1}}}\hspace{0.0002em}
\subfloat{\includegraphics[width=0.265\columnwidth]{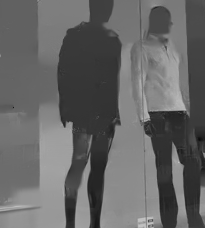}}\hspace{-0.06em}
\subfloat{\includegraphics[width=0.265\columnwidth]{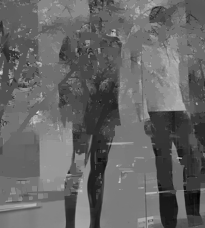}}\hspace{-0.06em}
\subfloat{\includegraphics[width=0.2135\columnwidth]{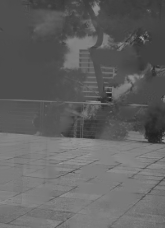}}\hspace{-0.06em}
\subfloat{\includegraphics[width=0.2135\columnwidth]{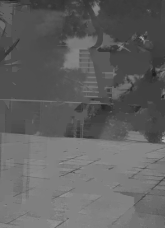}}\\[-2.5ex]
\subfloat{\rotatebox{90}{\footnotesize{\hspace{3.5em}Layer 2}}}\hspace{0.0002em}
\subfloat{\includegraphics[width=0.265\columnwidth]{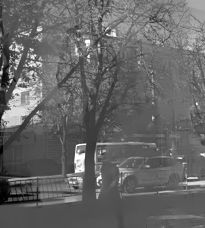}}\hspace{-0.06em}
\subfloat{\includegraphics[width=0.265\columnwidth]{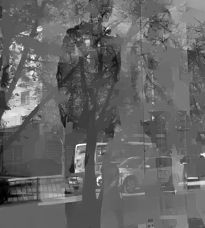}}\hspace{-0.06em}
\subfloat{\includegraphics[width=0.2135\columnwidth]{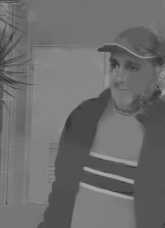}}\hspace{-0.06em}
\subfloat{\includegraphics[width=0.2135\columnwidth]{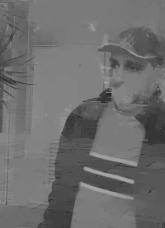}}
\end{tabular}}
\hspace{1.9em}GMM-C \hspace{3.9em} LW \hspace{2.9em} GMM-C \hspace{2.5em} LW \hspace{1em}
\caption{Decomposition results for real reflection images using the LW and GMM-C methods. Note that GMM-C annotations are not shown but that they were provided at the same locations as the LW annotations. Best viewed on screen.}
\label{fig:real_results}
\end{figure}

In figure \ref{fig:real_results} we show decompositions of real reflection images generated using the GMM-C and LW methods. In the GMM-C results we see successful separation of textured regions in one layer from overlapping edges and low frequency texture in the second layer  (see e.g. the mannequin shirt appearing in the left-most column). These textured regions are not successfully separated by the LW method.

Using non optimized code in Matlab on a standard PC the run time for optimizing the costs of the previous section on an image from BSDS300 is approximately two minutes. The code is available online at \href{http://github.com/ofersp/refsep}{github.com/ofersp/refsep}.

\section{Discussion}
\label{sec:discussion}
The single image reflection separation problem is inherently ill-posed and requires additional constraints provided by user annotations and a natural image prior. Previous work was based on a sparsity based prior and annotation mechanism which enabled semi-automatic separation of images in which both layers had little texture and the sparsity assumption was a good fit. However, for many real images, sparsity based methods cannot yield good separations even when a large fraction of the edges in the input image are correctly labeled. In this paper we have proposed a new user-assisted algorithm that is based on a much stronger prior of natural images - a GMM prior learnt from training examples. We have shown that this GMM, which has already been used successfully in image restoration tasks, can also be used to define a new annotation mechanism for user-assisted reflection separation. Our results show that high quality decompositions can be obtained with a relatively small amount of user interaction, even in the presence of significant texture.

\pagebreak
\bibliographystyle{IEEEbib}
\bibliography{refs}

\begin{thebibliography}{10}

\bibitem{ShihKrishnan}
YiChang Shih, Dilip Krishnan, Fredo Durand, and William~T Freeman,
\newblock ``Reflection removal using ghosting cues,''
\newblock in {\em Proceedings of the IEEE Conference on Computer Vision and
  Pattern Recognition}, 2015, pp. 3193--3201.

\bibitem{FaridAdelson}
Hany Farid and Edward~H Adelson,
\newblock ``Separating reflections and lighting using independent components
  analysis,''
\newblock in {\em Computer Vision and Pattern Recognition, 1999. IEEE Computer
  Society Conference on.} IEEE, 1999, vol.~1, pp. 262--267.

\bibitem{SchechnerShamir}
Yoav~Y Schechner, Joseph Shamir, and Nahum Kiryati,
\newblock ``Polarization-based decorrelation of transparent layers: The
  inclination angle of an invisible surface,''
\newblock in {\em Computer Vision, 1999. The Proceedings of the Seventh IEEE
  International Conference on}. IEEE, 1999, vol.~2, pp. 814--819.

\bibitem{szeliski2000layer}
Richard Szeliski, Shai Avidan, and P~Anandan,
\newblock ``Layer extraction from multiple images containing reflections and
  transparency,''
\newblock in {\em Computer Vision and Pattern Recognition, 2000. Proceedings.
  IEEE Conference on}. IEEE, 2000, vol.~1, pp. 246--253.

\bibitem{gai2012blind}
Kun Gai, Zhenwei Shi, and Changshui Zhang,
\newblock ``Blind separation of superimposed moving images using image
  statistics,''
\newblock {\em IEEE transactions on pattern analysis and machine intelligence},
  vol. 34, no. 1, pp. 19--32, 2012.

\bibitem{XueRubinstein}
Tianfan Xue, Michael Rubinstein, Ce~Liu, and William~T Freeman,
\newblock ``A computational approach for obstruction-free photography,''
\newblock {\em ACM Transactions on Graphics (TOG)}, vol. 34, no. 4, pp. 79,
  2015.

\bibitem{li2014single}
Yu~Li and Michael~S Brown,
\newblock ``Single image layer separation using relative smoothness,''
\newblock in {\em Proceedings of the IEEE Conference on Computer Vision and
  Pattern Recognition}, 2014, pp. 2752--2759.

\bibitem{yan2013separation}
Qing Yan, Yi~Xu, and Xiaokang Yang,
\newblock ``Separation of weak reflection from a single superimposed image
  using gradient profile sharpness,''
\newblock in {\em Circuits and Systems (ISCAS), 2013 IEEE International
  Symposium on}. IEEE, 2013, pp. 937--940.

\bibitem{LevinWeiss}
Anat Levin and Yair Weiss,
\newblock ``User assisted separation of reflections from a single image using a
  sparsity prior,''
\newblock {\em IEEE Transactions on Pattern Analysis and Machine Intelligence},
  vol. 29, no. 9, 2007.

\bibitem{MartinFTM01}
D.~Martin, C.~Fowlkes, D.~Tal, and J.~Malik,
\newblock ``A database of human segmented natural images and its application to
  evaluating segmentation algorithms and measuring ecological statistics,''
\newblock in {\em Proc. 8th Int'l Conf. Computer Vision}, July 2001, vol.~2,
  pp. 416--423.

\bibitem{Hyvarinen}
Aapo Hyv{\"a}rinen, Jarmo Hurri, and Patrick~O Hoyer,
\newblock {\em Natural Image Statistics: A probabilistic approach to early
  computational vision}, vol.~39,
\newblock Springer-Verlag New York Inc, 2009.

\bibitem{ZoranWeiss}
Daniel Zoran and Yair Weiss,
\newblock ``From learning models of natural image patches to whole image
  restoration,''
\newblock in {\em Computer Vision (ICCV), 2011 IEEE International Conference
  on}. IEEE, 2011, pp. 479--486.

\bibitem{Petersen}
Kaare~Brandt Petersen, Michael~Syskind Pedersen, et~al.,
\newblock ``The matrix cookbook,''
\newblock {\em Technical University of Denmark}, vol. 7, pp. 15, 2008.

\bibitem{LeviWeiss}
Effi Levi,
\newblock {\em Using natural image priors-maximizing or sampling?},
\newblock Ph.D. thesis, The Hebrew University of Jerusalem, 2009.

\end{thebibliography}

\end{document}